\documentclass[10pt, conference]{IEEEtran}

\usepackage{amsmath,amssymb,amsfonts}
\usepackage{graphicx}
\usepackage{pifont}

\usepackage{amsthm}
\usepackage{xcolor}
\usepackage[figuresright]{rotating}

\usepackage{graphicx}
\graphicspath{{/}{fig/}}

\usepackage{array}
\usepackage{textcomp}
\usepackage{xcolor}
\usepackage{multirow}
\usepackage{booktabs}

\usepackage{mathtools}
\usepackage{breqn}
\usepackage{float}

\usepackage{tikz,pgfplots}
\pgfplotsset{compat=1.7}
\usepgfplotslibrary{groupplots}

\usepackage{caption}
\usepackage{subcaption}

\usepackage{multirow,tabularx}
\usepackage{hyperref}
\usepackage{flushend}
\usepackage{algorithmic}
\usepackage[vlined, ruled, shortend]{algorithm2e}

\usepackage{adjustbox}
\usepackage{cleveref}

\newlength\figureheight
\newlength\figurewidth
\SetAlCapNameFnt{\footnotesize}
\SetAlCapFnt{\footnotesize}

\pgfplotsset{
    compat=1.11,
    legend image code/.code={
    \draw[mark repeat=2,mark phase=2]
        plot coordinates {
            (0cm,0cm)
            (0.15cm,0cm)        
            (0.3cm,0cm)         
        };%
    }
}

\title{
    Federated Learning for Vision-based Obstacle Avoidance in the Internet of Robotic Things
}

\author{
    \IEEEauthorblockN{
        \vspace{1em}
        Xianjia Yu\IEEEauthorrefmark{2},
        Jorge Pe\~na Queralta\IEEEauthorrefmark{2},
        Tomi Westerlund\IEEEauthorrefmark{2}
    }
    \IEEEauthorblockA{
        \normalsize
        \IEEEauthorrefmark{2}\href{https://tiers.utu.fi}{Turku Intelligent Embedded and Robotic Systems (TIERS) Lab, University of Turku, Finland}.\\
        Emails: \textsuperscript{1}\{xianjia.yu, jopequ, tovewe\}@utu.fi\\[+6pt]
    }
}

\begin{document}

\maketitle
\thispagestyle{empty}
\pagestyle{empty}



\begin{abstract}%
    \label{sec:abstract}%
    %
    Deep learning methods have revolutionized mobile robotics, from advanced perception models for an enhanced situational awareness to novel control approaches through reinforcement learning. This paper explores the potential of federated learning for distributed systems of mobile robots enabling collaboration on the Internet of Robotic Things. To demonstrate the effectiveness of such an approach, we deploy wheeled robots in different indoor environments. We analyze the performance of a federated learning approach and compare it to a traditional centralized training process with a priori aggregated data. We show the benefits of collaborative learning across heterogeneous environments and the potential for sim-to-real knowledge transfer. Our results demonstrate significant performance benefits of FL and sim-to-real transfer for vision-based navigation, in addition to the inherent privacy-preserving nature of FL by keeping computation at the edge. This is, to the best of our knowledge, the first work to leverage FL for vision-based navigation that also tests results in real-world settings.
    %
    %
\end{abstract}

\IEEEpeerreviewmaketitle


\section{Introduction}
\label{sec:introduction}

As ubiquitous autonomous mobile robots become increasingly interconnected, end-users and applications can benefit from distributed multi-robot systems~\cite{queralta2020collaborative}. Connected robots open a wide variety of opportunities within the Internet of Robotic Things (IoRT)~\cite{simoens2018internet}, specifically as mobile sensor networks capable of intelligent behavior and multi-modal data acquisition. We are particularly interested in exploring collaborative robot learning within the IoRT context~\cite{olcay2020collective, xianjia2021federated}, and studying the benefits that federated learning {FL} can bring to distributed robotic systems.

The relevance of AI and deep learning (DL) in robotic systems has increased significantly as DL methods enable higher degrees of situational awareness~\cite{pierson2017deep, karoly2020deep}. For a variety of tasks in robotics, mobile navigation, human-like walking, teaching through demonstration, and collaborative automation, to mention a few examples, DL or machine learning has attained state-of-the-art performance~\cite{pierson2017deep}. One of the areas where DL has had more impact is arguably computer vision, which has led us to put the focus on a relevant use-case for autonomous mobile robots: vision-based obstacle avoidance. Moreover, vision-based obstacle avoidance is instrumental not only for robot navigation but could also aid visually handicapped citizens~\cite{mendes2018assis}. Research has shown that DL can boost vision-based obstacle avoidance~\cite{gaya2016vision} and, especially when combined with specific technologies such as multi-view structure-from-motion, can reach a better degree of precision than other approaches~\cite{yang2017obstacle}.


\begin{figure}
    \centering
    \includegraphics[width=0.48\textwidth]{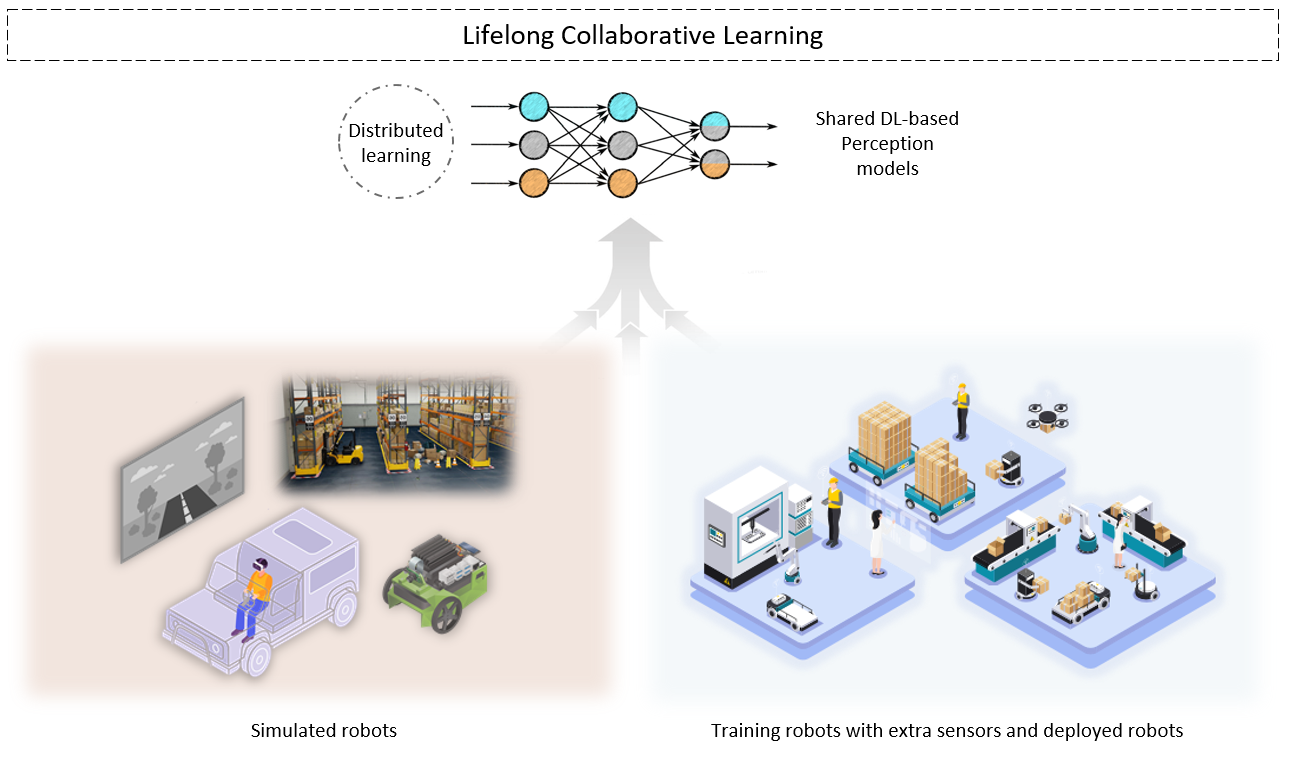}
    \caption{Conceptual illustration of federated lifelong learning with sim-to-real transfer as a deep learning-based perception model is trained in both a simulation and real robots, with potentially continuous updates.}
    \label{fig:fl_conceptual}
\end{figure}

In this manuscript, we focus on studying a federated learning approach for vision-based obstacle avoidance in distributed IoRT systems. A collective model built from and shared within a team of robots operating in different environments can bring numerous benefits, including more robust performance but more importantly higher readiness levels to operate in new environments or properly react to new situations. Collaborative multi-robot systems can be more efficient and have higher success rates in heterogeneous environments including unknown ones~\cite{olcay2020collective}.


Federated learning brings multiple benefits to collaborative learning in the IoRT. Compared to cloud-based centralized learning, it allows for optimization of networking resources and the preservation of data privacy by computing model updates directly at the edge. We propose a FL approach that combines data from both simulated and real robotic agents. A conceptual illustration of such system is illustrated in \cref{fig:fl_conceptual}. We study the performance of FL over centralized learning in the simulated and real worlds separately, analyze the improvements of merging data from heterogeneous scenarios, and finally the potential for sim-to-real knowledge transfer.


Partly owing to the benefits of sharing knowledge without transferring raw data, FL as a privacy-preserving distributed learning 
has been utilized in multiple domains in robotics and autonomous system~\cite{xianjia2021federated}. These domains include, but certainly are not limited to, navigation, cooperative SLAM based on visual-Lidar, trajectory forecasting, human-robot collaborative learning, and robot perception. To the best of our knowledge, however, there is no specific work related to applying FL in robotic vision-based obstacle avoidance in real-world scenarios.



In summary, in this work, we explore the potential for FL 
within hybrid teams of simulated and real robotic agents.
The main contributions of this work are the following. First, the design, implementation, deployment, and evaluation of a vision-based deep-learning approach to obstacle avoidance in mobile robots in heterogeneous simulated and real scenarios. We then evaluate the performance benefits of such an approach over offline learning or learning from more limited data sources. We put an emphasis on the benefits when robots are deployed in heterogeneous environments, showing that collaborative learning improves performance even for robots that do not change their environment. For this work, we deploy robots in highly photorealistic and physically-accurate virtual environments and study the ability of such a setup for sim-to-real transfer.


The remainder of this manuscript is organized as follows. Section II introduces related work in the relevant robot learning and federated learning literature. We then explain the methods and tools used for this work in Section III. Section IV reports the experimental results, while Section V concludes the work and lays down future research directions.



\section{Related Work} \label{sec:related_work}

This section introduces relevant examples in the literature in the areas of FL for robotics, DL for vision-based obstacle avoidance, and sim-to-real transfer. We also discuss different simulation environments that can be used for robot learning.

\subsection{Federated learning in robotics}

Robot collaboration has become vital as networked robots have become more ubiquitous~\cite{queralta2020collaborative}, and FL has stood as one of the best solutions from the point of view of data privacy, network resource management, and distributed edge computing~\cite{imteaj2021survey}. Other technologies, such as differential privacy, homomorphic encryption, and distributed ledger technologies (DLTs) have been used in the literature to improve FL from a systematic standpoint, making the collaborative learning process in a multi-robot system safer and privacy-preserving. FL offers potential in a variety of autonomy challenges and robotic subsystems, including cooperative SLAM, human-robot collaborative learning, and navigation, to name a few~\cite{xianjia2021federated}.

\subsection{Deep learning for vision-based obstacle avoidance}

Deep learning has been widely applied to robotic applications and has arguably revolutionized the role of AI in robotics~\cite{pierson2017deep, karoly2020deep, xianjia2022analyzing}. More specifically, it has gained popularity in vision-based obstacle avoidance because of the wide availability of vision sensors across platforms, and its suitability to different types of environments, including the absence of geometric models and substantial parameter tuning. In a recent relevant study~\cite{8917687}, the authors demonstrate the use of deep reinforcement learning (DRL) to perform obstacle avoidance via a monocular camera mounted on a UAV with minimal knowledge of the environment. In addition, DRL could perform more effectively even with irregular input by embedding the image's depth and semantic information. By adding noise to the depth information, the DRL model's resilience was increased to near-state-of-the-art levels in some unseen virtual and real-world circumstances~\cite{9636512}.


\subsection{Sim-to-real for robot learning}

Sim-to-real research has mostly focused on policies obtained through DRL~\cite{zhao2020sim}, and specifically in areas related to robotic manipulation~\cite{zhao2020towards}. Recent studies however have also applied DRL to vision-based obstacle avoidance in robot manipulators, demonstrating that learned models can be adapted efficiently to unseen scenes and unseen objects in the real world~\cite{zhang2020sim2real, zhang2021sim2real}. In this work, we take the sim-to-real analysis to mobile robots and explore the potential for collaborative learning of policies that can aid in robust autonomous navigation.

\subsection{Photorealistic simulation platforms}

Availability of data is one of the key challenges for the application of DL to robotics. Synthetic data and data from simulated environments have therefore played a key role in numerous studies. However, traditional robotics simulators such as Gazebo, widely used in the development of mobile robots, are incapable of generating realistic visuals. Therefore, while they are efficient for testing autonomy stacks based on other types of sensors such as lidars, which only need accurate geometric data, they are unsuitable for vision-based approaches and limit the sim-to-real knowledge transferability. A number of simulation platforms have emerged in recent years to address this issue. We have listed a relevant set of simulators in \cref{tab:3d_simulators}. Among these, Carla~\cite{dosovitskiy2017carla} has been widely used in research for self-driving cars, while AirSim~\cite{shah2018airsim} has been used in a wide variety of application scenarios. We have chosen for this study, however, the more recent NVIDIA Isaac Sim platform owing to the high-quality visuals but also tools that enable seamless generation of randomized environments for synthetic data acquisition. Randomization and the ability to alter the environments has been shown to be a key parameter to collaborative learning approaches~\cite{zhao2020towards, zhao2020ubiquitous}. An additional advantage is a common ecosystem of tools with the embedded NVIDIA Jetson platforms, the state-of-the-art in embedded computing for robots that need discrete GPUs for DL inference.


\begin{table}[t]
    \centering
    \caption{Comparison of existing 3D robotics simulators}
    \footnotesize
    \begin{tabular}{@{}lcccc@{}}
        \toprule
        \textbf{Simulator} & \textbf{Photorealistic} & \textbf{ROS} & \textbf{Multi-modal} \\
        \textbf{platform} & \textbf{visuals} & \textbf{support} & \textbf{sensors} \\
        \midrule
        Gazebo~\cite{koenig2004design} &   & \checkmark &  \checkmark \\
        VRKitchen~\cite{gao2019vrkitchen} & \checkmark &  &  \\
        MINOS~\cite{savva2017minos}  &  &  &  \\
        Gibson Env~\cite{xiazamirhe2018gibsonenv} & \checkmark & \checkmark &  &  \\
        Habitat~\cite{szot2021habitat} & \checkmark &  &  \\
        PreSim~\cite{yuan2021presim} & \checkmark &  &  \\
        Carla~\cite{dosovitskiy2017carla} & \checkmark & \checkmark & \checkmark \\
        AirSim~\cite{shah2018airsim} & \checkmark & \checkmark & \checkmark \\
        Nvidia Isaac~\cite{nvidia-sim} & \checkmark & \checkmark & \checkmark \\
        \bottomrule
    \end{tabular}
    \label{tab:3d_simulators}
\end{table}

\begin{figure*}
  \begin{subfigure}{0.33\textwidth}
    \includegraphics[width=\linewidth]{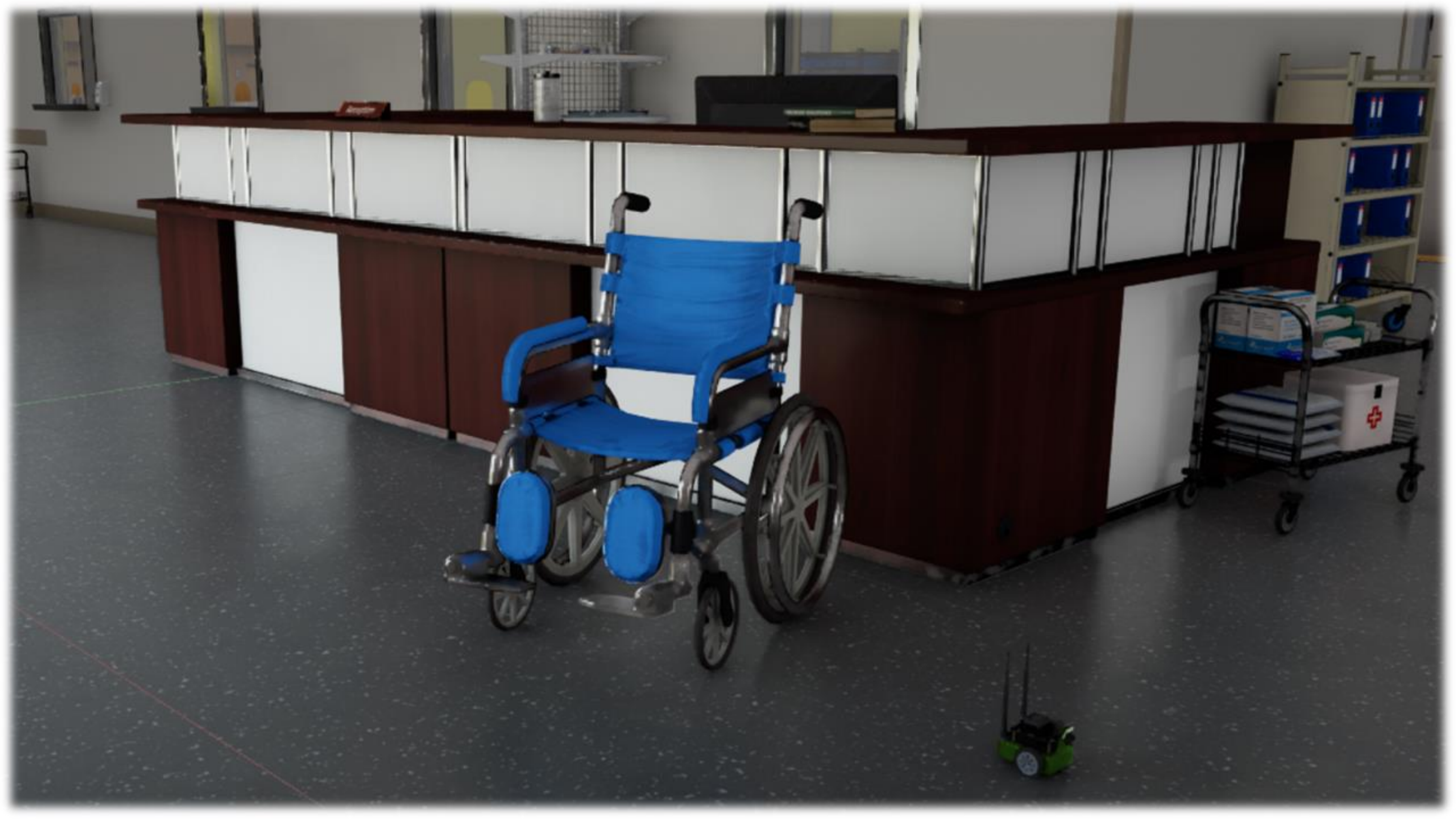}
    \caption{Hospital}
    \label{fig:hospital}
  \end{subfigure}%
  \hfill
  \begin{subfigure}{0.33\textwidth}
    \includegraphics[width=\linewidth]{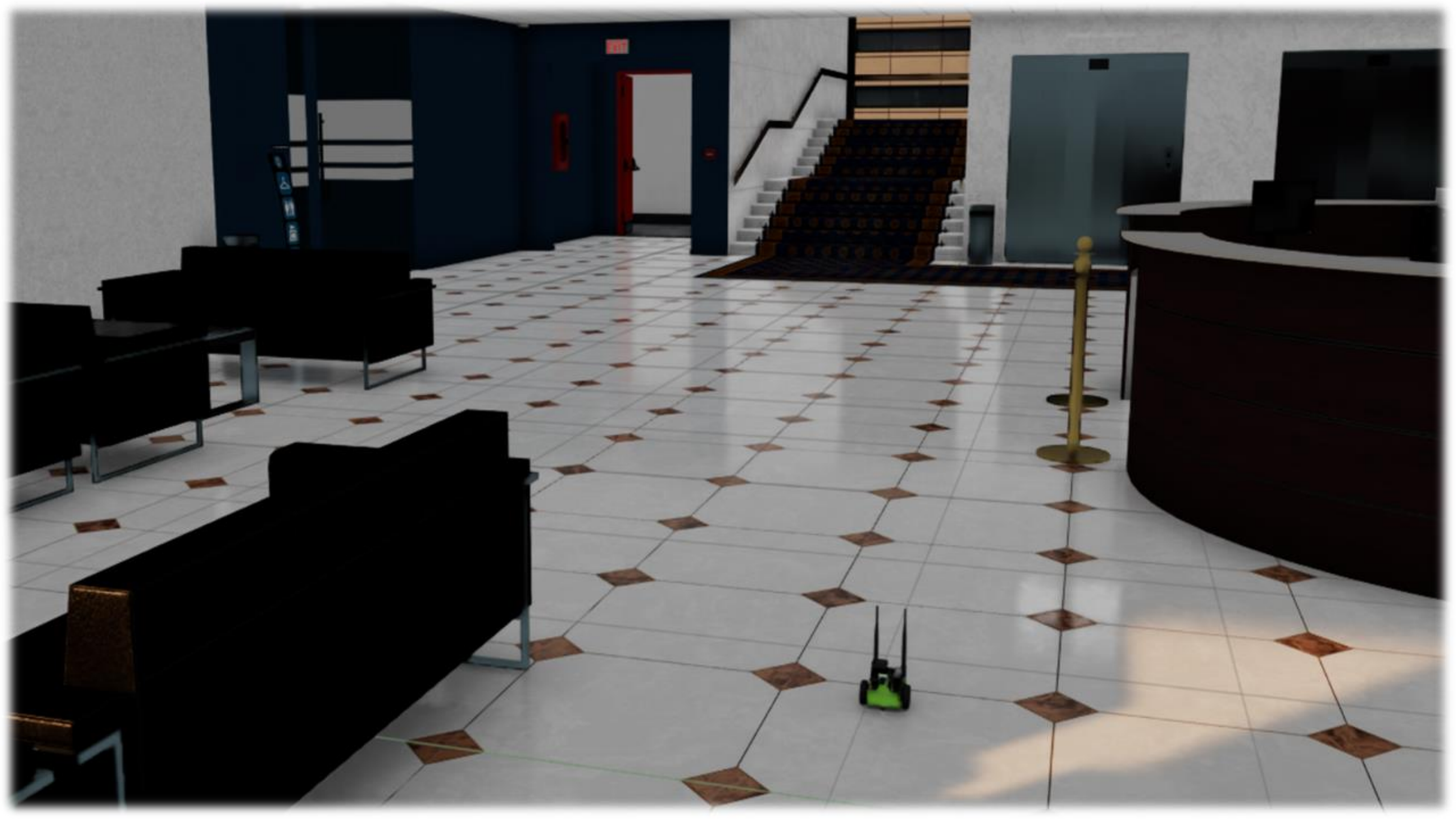}
    \caption{Office}
    \label{fig:office}
  \end{subfigure}%
  \hfill
  \begin{subfigure}{0.33\textwidth}
    \includegraphics[width=\linewidth]{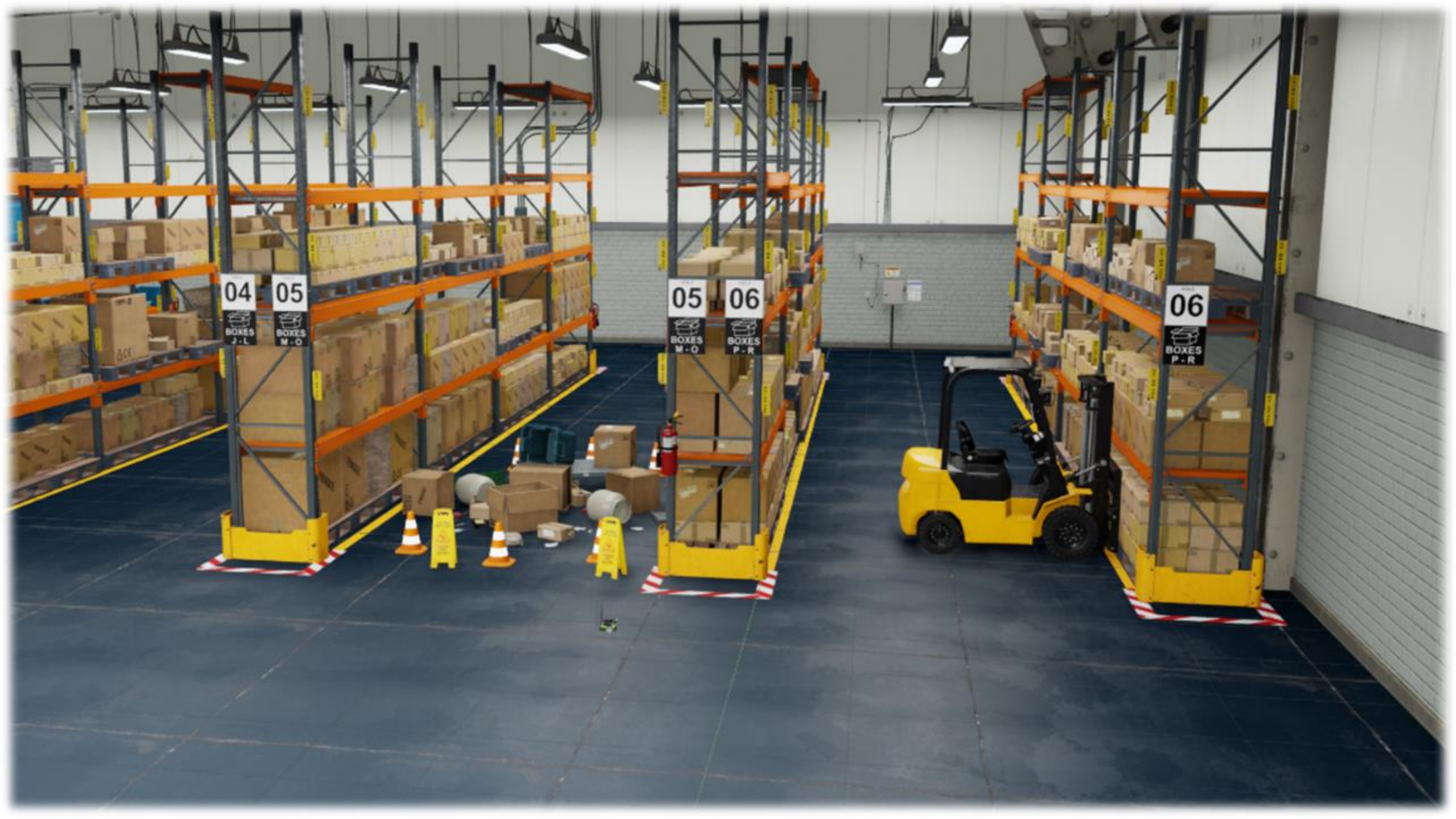}
    \caption{Warehouse}
    \label{fig:warehouse}
  \end{subfigure}
  \caption{Customized simulation environments in NVIDIA Isaac Simulator used in the experiments.}
  \label{fig:sim_environments}
\end{figure*}

\section{Methodology}\label{sec:methodology}

This section covers the different tools, simulation environments and robots utilized in the experiments. We describe the approaches to centralized and federated learning, and the deep learning models used for vision-based obstacle avoidance.

\subsection{Simulation settings}

As a platform to validate the proposed approach in simulated robots, we have used NVIDIA Isaac Sim, powered by Omniverse. Isaac Sim is a scalable robotics simulation application and synthetic data generation tool that enables the creation of photorealistic, physically accurate virtual environments for developing, testing, and managing AI-based robots~\cite{nvidia-sim}.


We have set up a series of simulation environments to gather data and validate the vision-based approach to obstacle avoidance. Three main environments are used to analyze the performance of the trained model, with a focus on heterogeneity of objects and backgrounds. The datasets used in this study include data from environments that replicate a hospital (see \cref{fig:hospital}), a office room (see \cref{fig:office}), and a warehouse (see \cref{fig:warehouse}).


To obtain sufficient data for model training, we used the NVIDIA Isaac Simulator's domain randomization (DR) and synthetic data recorder (SDR) features. By utilizing DR, we can select a specific object and randomly set its properties such as movement, rotation, light, and texture within a defined range. We can easily record the data generated by DR by using SDR. These operations were carried out in the three customized environments mentioned previously and pictured in \cref{fig:sim_environments}. The distribution of the datasets is shown in \cref{tab:data_distribution_sim}, with $\mathcal{S}_i$ representing the dataset associated to environment $i$.

\begin{table}
    \centering
    \caption{Distribution of simulation datasets}
    \label{tab:data_distribution_sim}
    \footnotesize
    \begin{tabular}{@{}lcccccccc@{}}
        \toprule
        & \multicolumn{2}{l}{\textbf{Hospital ($\mathcal{S}_1$)}} & \hspace{.23em} & \multicolumn{2}{c}{\textbf{Office ($\mathcal{S}_2$)}} & \hspace{.23em} & \multicolumn{2}{c}{\textbf{Warehouse ($\mathcal{S}_3$)}} \\[+.42em]
        & \textit{blocked} & \textit{free} & & \textit{blocked} & \textit{free} & & \textit{blocked} & \textit{free} \\
        \cmidrule{2-9}
        Prop. & 44\% & 56\% & & 64\% & 46\% & & 60\% & 40\% \\
        \cmidrule{2-3} \cmidrule{5-6} \cmidrule{8-9}
        Total & \multicolumn{2}{c}{27\%} & & \multicolumn{2}{c}{54\%} & & \multicolumn{2}{c}{19\%} \\ 
        \bottomrule
    \end{tabular}
\end{table}

%
%
%

\begin{table}
    \centering
    \caption{Distribution of real-world datasets}
    \label{tab:data_distribution_real}
    \footnotesize
    \begin{tabular}{@{}ccccccccc@{}}
        \toprule
        & \multicolumn{2}{l}{\textbf{Room 1 ($\mathcal{R}_1$)}} & \hspace{.23em} & \multicolumn{2}{c}{\textbf{Room 2 ($\mathcal{R}_2$)}} & \hspace{.23em} & \multicolumn{2}{c}{\textbf{Room 3 ($\mathcal{R}_3$)}} \\[+.42em]
        & \textit{blocked} & \textit{free} & & \textit{blocked} & \textit{free} & & \textit{blocked} & \textit{free} \\
        \cmidrule{2-9}
        Prop. & 40\% & 60\% & & 50\% & 50\% & & 50\% & 50\% \\
        \cmidrule{2-3} \cmidrule{5-6} \cmidrule{8-9}
        Total & \multicolumn{2}{c}{11\%} & & \multicolumn{2}{c}{44\%} & & \multicolumn{2}{c}{45\%} \\ 
        \bottomrule
    \end{tabular}
\end{table}


\subsection{Real-world experimental settings}

\begin{figure}[t]
        \centering
        \includegraphics[width=.5\textwidth]{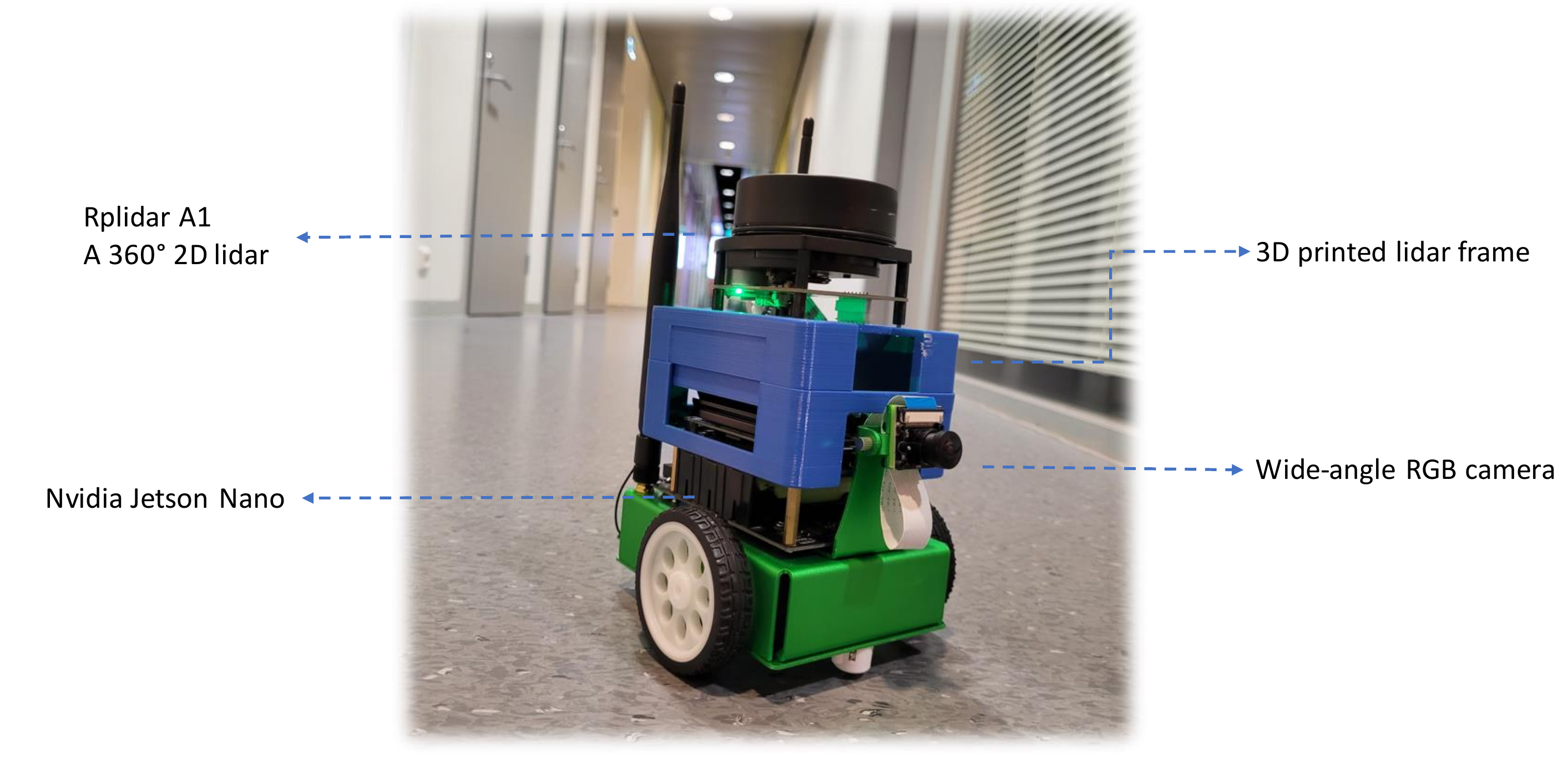}
        \caption{Customized Jetbot platform}
        \label{fig:jetbot}
\end{figure}

We utilize three mobile robots for real-world experiments and a local computing server for training the deep learning models. The platforms used in the experiments are three Jetbot robots from Waveshare (depicted in \cref{fig:jetbot}), equipped by default with a wide-angle lens RGB camera and an embedded NVIDIA Jetson Nano development kit. In addition, a Rplidar A1 2D rangefinder has been installed on a 3D printed frame for automated data labeling when real-world data is collected. The local edge server used for training the obstacle avoidance models is equipped with an 8-core Intel i7-9700K processor, 64\, GB of memory, and an NVIDIA GeForce RTX 2080 Ti discrete graphics card.



We deployed the mobile robots in three different indoor environments (office spaces, hallways and laboratory environments) to validate the obstacle avoidance policies trained with the different approaches. The three rooms vary in terms of objects present as obstacles, material texture, layout, and style. 
The distribution of the images acquired by the three robots is shown in \cref{tab:data_distribution_real}, where $\mathcal{R}_{i,\:i\in\{0,1,2\}}$ represent each of the acquired datasets for the respective real-world environments. We intentionally imbalance the data in order to imitate a situation in which robots would not be able to collect data equally across different environments and operational conditions. We also do this to evaluate whether there is a performance impact in environments based on the amount of collected data. As such, \textit{Room 1} only accounts for 11\% of the total amount of collected images.


\subsection{Vision-based obstacle avoidance models}
\label{subsec:voa_training}
In order to achieve vision-based obstacle avoidance for the operation of mobile robots, we have chosen to train a generic DL model to assist robots in discriminating between different types of obstacles across heterogeneous environments. This approach comes in contrast to other options, including the detection of individual objects or semantic segmentation (e.g., for segmenting free floor from objects and walls). The selected approach enables us to focus on analyzing the performance of a federated learning approach and the ability for sim-to-real transfer rather than on the design of a specific obstacle avoidance strategy, which is the main objective of this study

More precisely, we utilized a deep convolutional neural network (CNN) to carry out a vision-based obstacle classifier for two only two classes, that define whether the environment ahead is \textit{blocked} or \textit{free} for the robot to navigate. Owing to the relative low level of complexity of the classifier and the limited size of the collected datasets, we have selected the AlexNet~\cite{krizhevsky2012imagenet} architecture as appropriate for such binary classification task. AlexNet has been established as the precedent for deep CNN as one of the most widely used backbones for executing various tasks across multiple domains.

We train different models for each separate dataset with both simulated and real data, as well as combinations of these. The models are trained using two approaches: a centralized learning approach that aggregates data from all robots in the local edge server and trains them at once; and a federated learning approach that only fuses the individual models trained in each of the different scenarios.




\section{Experimental Results}

Through this section, we report the experimental results obtained with data from both simulated and real robots. We show first the performance of the different approaches, with the latter part shifting towards the potential for sim-to-real knowledge transferability.

\subsection{Centralized training vs. federated learning}

\begin{figure*}[t]
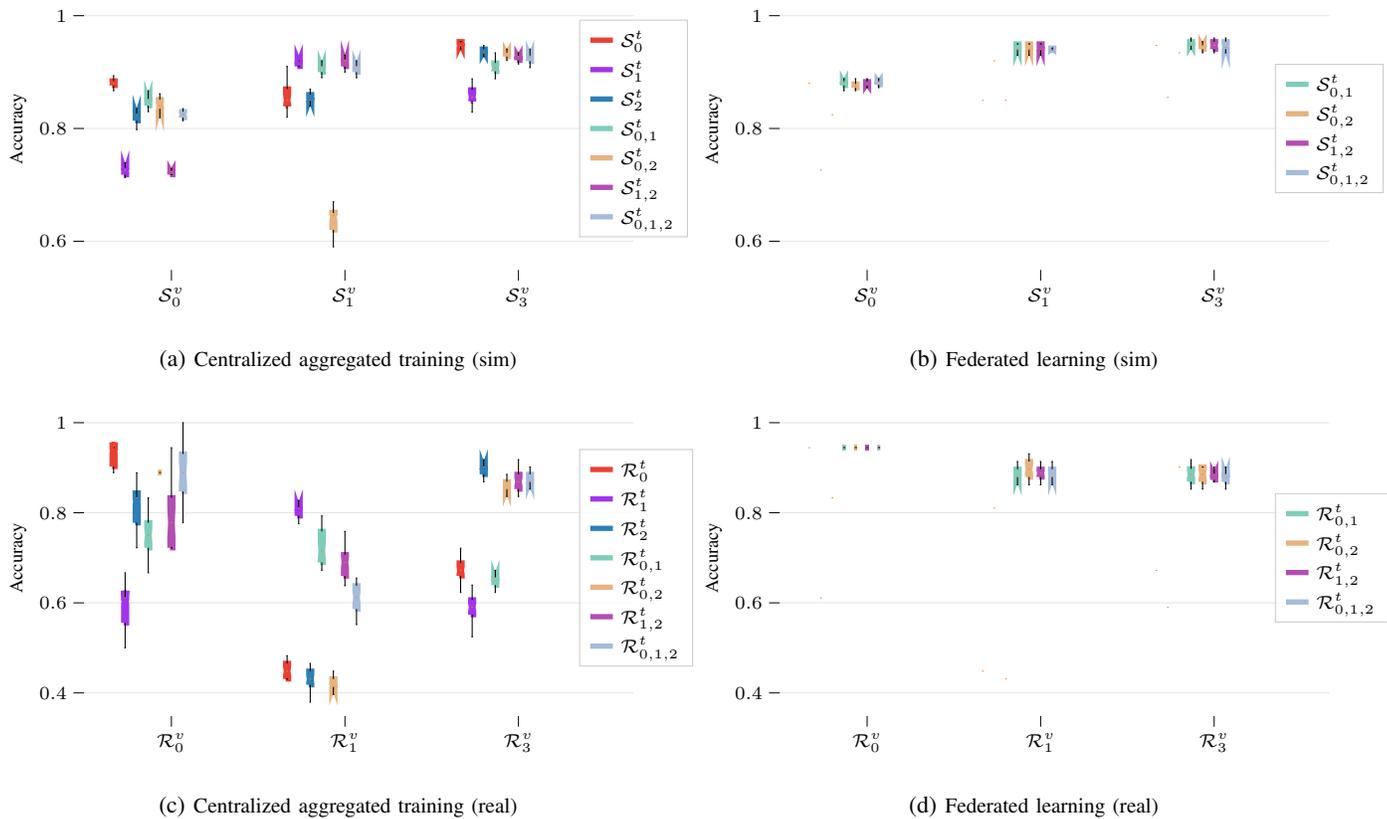

    \centering
    \begin{subfigure}{.49\textwidth}
        \centering
        \setlength{\figurewidth}{\textwidth}
        \setlength{\figureheight}{.6\textwidth}
        \scriptsize{\input{tex/sim_mt_acc}}
        \caption{Centralized aggregated training (sim)}
        \label{fig:sim_acc_mixed}
    \end{subfigure}
    \hfill
    \begin{subfigure}{.49\textwidth}
      \centering
        \setlength{\figurewidth}{\textwidth}
        \setlength{\figureheight}{.6\textwidth}
        \scriptsize{\input{tex/sim_mf_acc}}
        \caption{Federated learning (sim)}
      \label{fig:sim_acc_fused}
    \end{subfigure}
    
    \vspace{1em}
    
    \begin{subfigure}{.49\linewidth}
        \centering
        \setlength{\figurewidth}{\textwidth}
        \setlength{\figureheight}{.65\textwidth}
        \scriptsize{\input{tex/real_mt_acc}}
        \caption{Centralized aggregated training (real)}
        \label{fig:real_acc_mixed}
    \end{subfigure}
    \hfill
    \begin{subfigure}{.49\linewidth}
        \centering
        \setlength{\figurewidth}{\textwidth}
        \setlength{\figureheight}{.65\textwidth}
        \scriptsize{\input{tex/real_mf_acc}}
        \caption{Federated learning (real)}
        \label{fig:real_acc_fused}
    \end{subfigure}
    \caption{Accuracy of the different models obtained through centralized learning with aggregated data or federated learning with fused local models. These results are trained ($^t$) and validated ($^v$) with respective simulation datasets ($\mathcal{S}_i^t$, $\mathcal{S}_i^v$) and real datasets ($\mathcal{R}_i^t$, $\mathcal{R}_i^v$) independently.}
    \label{fig:accuracy_results}
\end{figure*}

The first objective of our experiments is to analyze the performance improvements that a federated learning approach brings over a centralized training with traditional data aggregation. To do this, we used the data we collected in the simulated hospital ($\mathcal{S}_0$), office ($\mathcal{S}_1$), and warehouse ($\mathcal{S}_2$) to train our model on each dataset and all possible combinations of two or three of the datasets. Equivalently for the federated learning approach, we run different training rounds in which we simulate that a different subset of robots is collaboratively learning without sharing any actual raw data. In this approach, only the models are fused and a common model updated iteratively. \cref{fig:sim_acc_mixed} and \cref{fig:sim_acc_fused} report the accuracy of the different models for the centralized and federated approaches, respectively. For the FL results, the training happens only with combination of datasets from different environments.

The accuracy of models trained with real-world data is then shown in \cref{fig:real_acc_mixed} and \cref{fig:real_acc_fused}. The data has been obtained with Jetbots navigating in three different office and laboratory indoor environments ($\mathcal{R}_{i,\:i\in{0,1,2}}$).

In addition to the accuracy, we also calculate the area under the ROC curve (AUC) for each of the scenarios where training is carried out through either the centralized or federated approaches. The results are reported in \cref{tab:auc_values} This metric gives a better understanding of the reliability of the models. In this particular application scenario of robotic navigation, there is indeed a disparity in the cost of false negatives over false positives in terms of the robot's integrity. However, from the point of view of performance, false positives can degrade significantly the navigation speed and time, while low-frequency collisions can be mitigated with, e.g., bumper sensors. Therefore, the classification-threshold invariance of the AUC metric is relevant to this use case.

Through both the accuracy and AUC results, we see that there is a clear improvement when the models and not the data are aggregated. In addition to the better navigation results, this also brings other advantages. First, we optimize the networking resources, allowing for intermittent connectivity and potentially lower bandwidth requirements when the size of the data in training batches is significantly smaller than the models. Moreover, this allows for privacy-preserving collaboration between different end-users or robot operators, as the raw data does not need to be exposed to a central authority or service.

\begin{table*}
    \centering
    \caption{Area under ROC curve (AUC) values for the aggregated centralized learning and federated learning approaches.}
    \label{tab:auc_values}
    \small
    \begin{tabular}{@{}cclcccccccccccc@{}}
        \toprule \\[-.85em]
        \hspace{.42em} & \hspace{.42em} & & \multicolumn{12}{c}{Training datasets} \\[+.25em]
        \cmidrule{4-15} \\[-.85em]
        & & & \multicolumn{7}{c}{Centralized learning with aggregated data} &  & \multicolumn{4}{c}{Federated learning} \\[+.25em]
        \cmidrule{4-10} \cmidrule{12-15} \\[-.55em]
        \multicolumn{1}{c}{\multirow{12}{*}{\rotatebox[origin=l]{90}{\hspace{1em} Validation datasets \hspace{1em}}}} & & & $\mathcal{S}_0^t$ & $\mathcal{S}_{1}^t$ & $\mathcal{S}_{2}^t$ & $\mathcal{S}_{0,1}^t$ & $\mathcal{S}_{0,2}^t$ & $\mathcal{S}_{1,2}^t$ & $\mathcal{S}_{1,2,3}^t$ & & $\mathcal{S}_{0,1}^t$ & $\mathcal{S}_{0,2}^t$ & $\mathcal{S}_{1,2}^t$ & $\mathcal{S}_{1,2,3}^t$ \\[+0.6em]
        & \parbox[t]{1mm}{\multirow{3}{*}{\rotatebox[origin=c]{90}{Sim}}}
        & \textbf{$\mathcal{S}_{0}^v$} & 0.28 & 0.56 & 0.56 & 0.33 & 0.50 & 0.71 & 0.52 & & 0\textbf{.85} & \textbf{0.85} & \textbf{0.85} & \textbf{0.85} \\
        & & \textbf{$\mathcal{S}_{1}^v$} & 0.33 & 0.50 & 0.75 & 0.33 & 0.50 & 0.75 & 0.60 & & 0.94 & 0.93 & 0.93 & \textbf{0.95} \\
        & & \textbf{$\mathcal{S}_{2}^v$} & 0.43 & 0.94 & 0.46 & 0.42 & 0.50 & 0.23 & 0.62 & & \textbf{0.96} & 0.95 & 0.95 & 0.95 \\
        \\[-0.4em]
        
        & & & $\mathcal{R}_0^t$ & $\mathcal{R}_{1}^t$ & $\mathcal{R}_{2}^t$ & $\mathcal{R}_{0,1}^t$ & $\mathcal{R}_{0,2}^t$ & $\mathcal{R}_{1,2}^t$ & $\mathcal{R}_{1,2,3}^t$ & & $\mathcal{R}_{0,1}^t$ & $\mathcal{R}_{0,2}^t$ & $\mathcal{R}_{1,2}^t$ & $\mathcal{R}_{1,2,3}^t$ \\[+0.6em]
        & \parbox[t]{1mm}{\multirow{3}{*}{\rotatebox[origin=c]{90}{Real}}}
        & $\mathcal{R}_0^v$ & 0.63 & 0.58 & 0.42 & 0.75 & 0.63 & 0.08 & 0.58 & & \textbf{0.88} & \textbf{0.88} & \textbf{0.88} &\textbf{ 0.88} \\
        & & $\mathcal{R}_1^v$ & 0.31 & 0.66 & 0.60 & 0.35 & 0.25 & 0.50 & 0.70 & & 0.83 & \textbf{0.85} & \textbf{0.85} & \textbf{0.85} \\
        & & $\mathcal{R}_2^v$ & 0.69 & 0.57 & 0.87 & 0.46 & 0.51 & 0.48 & 0.56 & & 0.90 & \textbf{0.92} & 0.90 & \textbf{0.92} \\
        \bottomrule
    \end{tabular}
\end{table*}

\subsection{Sim-to-real performance evaluation}

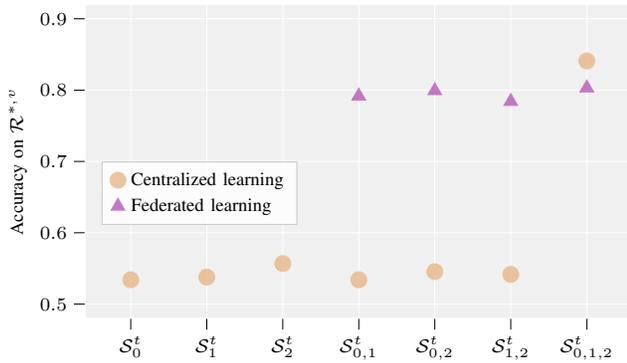
\begin{figure}[t]
    \centering
    \setlength{\figurewidth}{\linewidth}
    \setlength{\figureheight}{.65\linewidth}
    \scriptsize{
    
    
    

\begin{tikzpicture}

\definecolor{color0}{rgb}{0.9,0.7,0.5}
\definecolor{color1}{rgb}{0.7,0.3,0.7}

\begin{axis}[
    height=\figureheight,
    width=\figurewidth,
    legend cell align={left},
    legend style={
      fill opacity=0.8,
      draw opacity=1,
      text opacity=1,
      at={(0.03,0.5)},
      anchor=north west,
      draw=white!80!black
    },
    axis line style={white},
    axis background/.style={fill=white!94!black},
    tick align=outside,
    tick pos=left,
    x grid style={white},
    xtick style={color=black},
    y grid style={white},
    ymajorgrids,
    ytick style={color=black},
    scaled y ticks = false,
    %
    %
    %
    %
    %
    %
    xmajorgrids,
    xmin=0.4, xmax=7.6,
    xtick style={color=black},
    xtick={1,2,3,4,5,6,7},
    xticklabels={$\mathcal{S}^t_0$,$\mathcal{S}^t_1$,$\mathcal{S}^t_2$,$\mathcal{S}^t_{0,1}$,$\mathcal{S}^t_{0,2}$,$\mathcal{S}^t_{1,2}$,$\mathcal{S}^t_{0,1,2}$},
    ylabel={Accuracy on $\mathcal{R}^{*,v}$},
    ymajorgrids,
    ymin=0.48, ymax=0.92,
    ytick style={color=black},
    ytick={0.5,0.6,0.7,0.8,0.9},
]
\addplot [only marks, fill opacity=0.75, draw opacity=0.72, color0, mark=*, mark size=3, mark options={solid}]
table {%
1 0.534090936183929
2 0.537878811359406
3 0.556818187236786
4 0.534090936183929
5 0.545454561710358
6 0.541666686534882
7 0.840909063816071
};
\addlegendentry{Centralized learning}
\addplot [only marks, fill opacity=0.75, draw opacity=0.72, color1, mark=triangle*, mark size=3, mark options={solid}]
table {%
4 0.791666686534882
5 0.799242436885834
6 0.784090936183929
7 0.80303031206131
};
\addlegendentry{Federated learning}
\end{axis}

\end{tikzpicture}}
    \caption{Sim-to-real accuracy of simulation-trained models validated on an independent real-world navigation dataset.}
    \label{fig:sim2real_val}
\end{figure}

In the last part of our experiments, we evaluate the ability of both the centralized and federated learning approaches to transfer knowledge from simulation environments to the real world. To do this, we rely on the same simulation environments but introduce an independent data validation set ($\mathcal{R}^*$) from a navigation mission across different types of indoor spaces. The accuracy for each of the trained models is shown in \cref{fig:sim2real_val}. We can observe that relatively low performance is achieved with either approach when only one of the simulation environments is used for training the models. This may result from overfitting the model to non-realistic features in the simulated worlds. However, when heterogeneous data is introduced in training, the federated learning approach significantly improves. Our results also show that only when aggregating data from all three simulation environments the centralized learning approach is able to improve the performance to the level of federated learning. The specific reason behind this behaviour requires further study and will be the object of future research.


\section{Discussion and Analysis}
\label{sec:conclusion}

We have presented a federated learning approach for vision-based obstacle avoidance in mobile robots that leverages data from both simulated agents and real robots with additional sensors. We have shown that interconnected robots relying on deep learning for vision-based navigation can aid each other without sharing raw data. Specifically, we show how training the same model with data from heterogeneous environments improves performance across the simulated and real worlds. More importantly, the performance improvements are better when the models are trained through a federated learning approach compared to centralized learning. In addition to the application-specific improvements, the federated learning approach brings inherent benefits in terms of communication optimization and preservation of data privacy, enabling collaboration across organizations or users. Finally, we have shown that the presented approach is able to transfer knowledge from simulation to reality effectively.

Owing to the potential for sim-to-real transfer and accounting for the better performance of federated over centralized learning, future work will be directed towards lifelong FL through a combination of simulated and real agents. Additionally, we will further explore the reasons behind the differences in performance across the approaches presented in this manuscript.



\section*{Acknowledgment}

This research work is supported by the Academy of Finland's AutoSOS project (Grant No. 328755) and RoboMesh project (Grant No. 336061).

\bibliographystyle{unsrt}
\bibliography{bibliography}

\end{document}